\documentclass[runningheads]{llncs}
\usepackage{multirow}

\usepackage[mathscr]{eucal} 
\usepackage{amsbsy} 
\usepackage{times}
\usepackage{latexsym}
\usepackage{float} 
\usepackage{cite}

\usepackage[T1]{fontenc}
\usepackage{wrapfig}
\usepackage{booktabs}
\usepackage{xcolor}
\usepackage[colorlinks=true,linkcolor=black,citecolor=darkblue,urlcolor=darkblue]{hyperref} 

\definecolor{darkblue}{RGB}{0,0,139}

\usepackage[utf8]{inputenc}

\usepackage{inconsolata}

\newcommand{\tensor}[1]{\boldsymbol{\mathscr{#1}}} 

\usepackage{graphicx}
\usepackage{amsmath}
\begin{document}
\title{TRAWL: Tensor Reduced and Approximated Weights for Large Language Models}
%
%

\author{
  Yiran Luo\inst{1}\thanks{Yiran Luo and Het Patel contributed equally to this work.} \and
  Het Patel\inst{1} \and
  Yu Fu\inst{1} \and
  Dawon Ahn\inst{1} \and
  Jia Chen\inst{2} \and
  Yue Dong\inst{1} \and
  Evangelos E. Papalexakis\inst{1}
}

\authorrunning{Y. Luo et al.}

\institute{
  Department of Computer Science and Engineering, UC Riverside, USA \\
  \email{\{yluo147, hpate061, yfu093, dahn017, yued, epapalex\}@ucr.edu}
  \and
  Department of Electrical and Computer Engineering, UC Riverside, USA \\
  \email{jiac@ucr.edu}
}


\maketitle              
\begin{abstract}
Recent research has shown that factorizing large-scale language models for inference is an effective approach to improving model efficiency, significantly reducing model weights with minimal impact on performance. Interestingly, factorization can sometimes even improve accuracy by removing the noise that accumulates during training, particularly through matrix decompositions. However, recent work has primarily focused on single-matrix decompositions or lower precision techniques, which may fail to fully capture structural patterns. To address these limitations, we introduce \textbf{TRAWL} (Tensor Reduced and Approximated Weights for Large Language Models), a technique that applies tensor decomposition across \textit{multiple weight matrices} to effectively denoise LLMs by capturing both global and local structural patterns. Our experiments show that TRAWL improves the model performance by up to \textbf{16\%} over baseline models on benchmark datasets, without requiring additional data, training, or fine-tuning.
\end{abstract}
\section{Introduction}

The success of Transformer-based models like the GPT series \cite{radford2018improving, radford2019language, brown2020language}, and LLaMA \cite{touvron2023llama} is often attributed to their large scale \cite{kaplan2020scaling}. While the large number of parameters allows these models to capture complex patterns, training and deploying them requires significant computational resources and energy \cite{samsi2023words, zhao2022green}, making them impractical for many real-world applications. 

To address these limitations, recent studies have explored post-training compression techniques, showing minimal impact on inference performance \cite{han2015learning, sharma2023truth, saha2024compressing}. Interestingly, some research suggests that certain factorization methods can reduce computation time while also enhancing performance. For instance, LASER \cite{sharma2023truth} applies Singular Value Decomposition (SVD) on weight matrices for model compression, demonstrating that removing noise introduced during training can improve performance. CALDERA \cite{saha2024compressing} builds on LASER by further compressing weight matrices into lower precision formats, achieving significant size reductions and lower computational costs. However, both methods operate on individual matrices, potentially missing broader structural patterns across the model.

In this paper, we introduce TRAWL (Tensor Reduced and Approximated Weights for LLMs), a novel approach that extends denoising by leveraging tensor decomposition across \textbf{multiple weight matrices}. Unlike methods that focus on individual matrices, TRAWL stacks multiple weight matrices into higher-order tensors, enabling more effective exploitation of the model's inherent structures.

Our experiments on two Transformer-based models—RoBERTa \cite{liu2019roberta} and GPT-J \cite{gpt-j} across three datasets (BigBench WikiQA \cite{srivastava2022beyond}, BiosProfession \cite{10.1145/3287560.3287572}, and HotpotQA \cite{yang2018hotpotqadatasetdiverseexplainable}) demonstrate that TRAWL consistently outperforms single matrix decomposition approaches, achieving accuracy improvements of up to 16\%. By leveraging advanced tensor decomposition, TRAWL effectively reduces noise introduced during training and enhances performance without requiring additional training or fine-tuning. These results highlight TRAWL as a practical and efficient optimization technique for LLMs, delivering significant accuracy gains over baseline models.

Our key contributions are:
\begin{itemize}
    \item  We propose a post-training compression approach that uses tensor decomposition across multiple weight matrices, enabling more effective noise reduction and rank compression of model weights.
    
    \item 
We conducted extensive experiments on two models across three datasets, showing that TRAWL improves accuracy and generalization over existing methods without additional training or fine-tuning, making it highly practical for real-world applications. All code from this study is publicly available to support transparency and further research at \href{https://github.com/HettyPatel/TRAWL}{GitHub Link}.

\end{itemize}

\section{Related Work}
Recent efforts in neural network compression have focused on matrix or tensor decomposition techniques to reduce computational or storage requirements while maintaining or even improving performance.

LAyer SElective Rank reduction (LASER)  \cite{sharma2023truth} and Calibration Aware Low-Precision DEcomposition with Low-Rank Adaptation (CALDERA) \cite{saha2024compressing} are two representative methods which utilize matrix decomposition and low-rank approximation on LLM weights. LASER reduces the rank of weight matrices with singular value decomposition (SVD), enhancing performance by reducing noise from higher-order components. CALDERA decomposes a weight matrix into a low-precision backbone matrix and low-rank factors, reducing the model's memory footprint while maintaining performance.

A recent study
\cite{yu2023compressing} challenges the assumption that the model weights in transformers are low rank. Their study, titled "Compressing Transformers: Features Are Low-Rank, but Weights Are Not!" demonstrates that while the features (i.e., activations) of transformers exhibit low-rank characteristics, the model weights do not. This finding is particularly significant, as it contrasts with the foundation of several compression techniques, such as LASER, which assume that low-rank weight approximations can be beneficial. The authors propose an adaptive approach to compress model features instead of weights, achieving substantial parameter reductions with minimal accuracy loss using a few-shot unsupervised method. 

We acknowledge this distinction, but also observe that, in some cases, low-rank approximations of weights, although not necessarily as low-rank as one would normally expect from the popular use of the term, can still offer performance improvements with approximation ranks that are still lower than the original dimensionality of the weights. This nuanced understanding aligns with our results, where we find that certain configurations and ranks, even if not conventionally low (low compared to the dimensionality of weights), provide significant improvements. Thus, while our approach involves weight matrix decomposition, it does so with a flexible consideration of rank, corroborating, and extending the findings of the "Compressing Transformer" study. 

Tensor train decomposition (TTD) \cite{oseledets2011tensor} decomposes weight tensors into low-rank chains, effectively reducing parameters and computational complexity. Similarly, tensor ring decomposition (TRD) \cite{zhao2016tensor} arranges tensors in a ring structure for further model size reduction.

Lebedev et al. \cite{lebedev2014speeding} proposed a two-step approach for speeding up the training for large convolutional neural networks (CNNs) using tensor decomposition and discriminative fine-tuning, which achieves significant speed-ups in training with minor accuracy drops.

Tensor Contraction Layers (TCLs) \cite{kossaifi2017tensor} incorporated tensor contractions as end-to-end trainable neural network layers, leading to significant model compression with minimal impact on accuracy and, in some cases, improved performance.

Low-Rank Adaptation (LoRA) \cite{hu2021lora} focuses on introducing trainable low-rank matrices into the transformer architecture for efficient fine-tuning, significantly reducing trainable parameters, and lowering GPU memory requirements without sacrificing model performance. Similarly, Low-Rank Economic Tensor-Train Adaptation (LoRETTA) \cite{yang2024loretta} employs tensor-train decomposition to achieve significant parameter reduction while maintaining performance.

Our method, TRAWL, builds on these approaches by fully exploiting tensor structures in LLMs, achieving enhanced performance without additional training or fine-tuning. To our knowledge, this study is the first to explore the performance benefits of tensor factorization in LLMs in a setting where fine-tuning does not occur.





\section{Methodology}

\begin{figure}[t]
  \centering
  \includegraphics[width=0.65\textwidth]{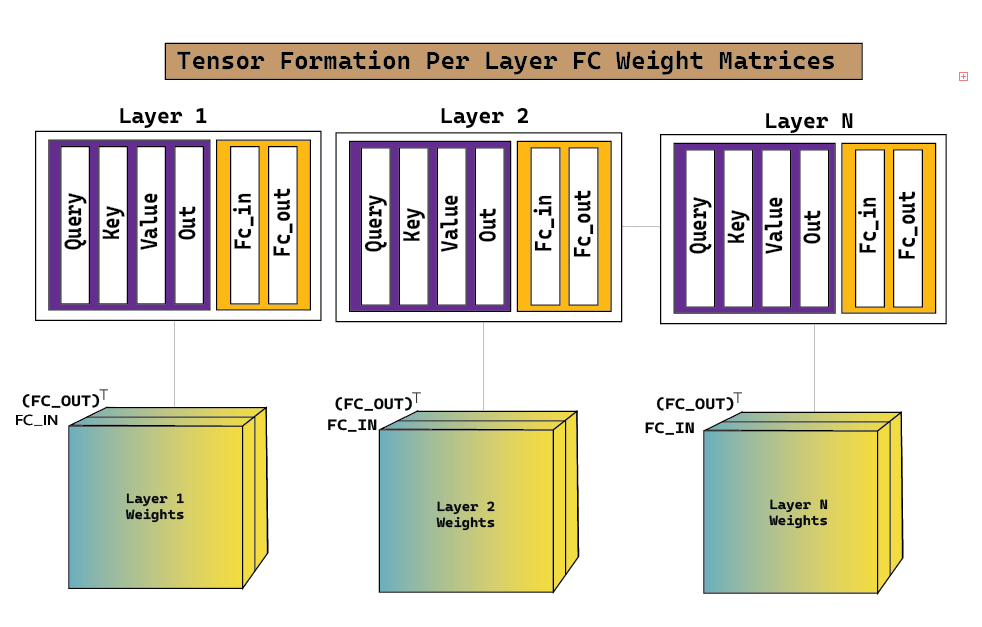}
  \caption{Tensor formulation by stacking QKVO or FC weights from a single layer. A 3-mode tensor is created for each layer, and tensor decomposition is applied.}
  \label{fig:TensorPerLayer}
\end{figure}

This section provides background information on tensor decomposition and describes our proposed method.

\subsection{Tensor Decomposition}

Tensor decomposition is a technique for analyzing multidimensional data by breaking a tensor into lower dimensional components \cite{sidiropoulos2017tensor}. A tensor, essentially a multidimensional array, can be decomposed using various tensor decomposition methods. In this paper, we focus on two: CANDECOMP/PARAFAC (CP) decomposition and Tucker decomposition.

CP decomposition represents a tensor as a sum of rank-one tensors. For a tensor $\mathcal{W}$ of order three, this is expressed as:

\begin{equation}
    \tensor{W} \approx \hat{\tensor{W}} = \sum_{r=1}^{R} \mathbf{a}_r \circ \mathbf{b}_r \circ \mathbf{c}_r
\end{equation}

where $\mathbf{a}_r$, $\mathbf{b}_r$, and $\mathbf{c}_r$ are factor vectors, $\circ$ denotes the outer product, and $R$ is the rank. $\hat{\tensor{W}}$ is the low-rank approximation of $\tensor{W}$ and it represents the original tensor as a sum of a few rank-one tensors. This reduces storage and computation while preserving essential information.

In the case of Tucker decomposition, $\tensor{W}$ is approximated as a core tensor $\tensor{G}$ multiplied by factor matrices $\mathbf{A}$, $\mathbf{B}$, and $\mathbf{C}$ along each mode:

\begin{equation} 
\tensor{W} \approx \hat{\tensor{W}} = \tensor{G} \times_1 \mathbf{A} \times_2 \mathbf{B} \times_3 \mathbf{C} 
\end{equation}
where $\times_n$ denotes the $n$-mode product and essentially multiplies all the slices of the $n$-th mode of the tensor with the corresponding factor matrix.

We use least squares fitting via TensorLy \cite{tensorly}. The number of rank-one components, $R$, is treated as a hyperparameter, and we explore a wide range of values. Finding the optimal $R$ remains a challenging problem \cite{shiao2022frappe}, beyond the scope of this study.

\subsection{Proposed Approach}
\label{sec:modes}

\begin{table*}[t!]
\centering
\resizebox{\textwidth}{!}{
\begin{tabular}{l|l|c|c|c|c|c|c}
\toprule
\multirow{3}{*}{\textbf{Model}} & \multirow{3}{*}{\textbf{Approach}} & \multicolumn{2}{c|}{\textbf{BigBench WikiQA}} & \multicolumn{2}{c|}{\textbf{BiosProfession}} & \multicolumn{2}{c}{\textbf{HotpotQA}} \\ \cmidrule{3-8} 
                               &                         & Accuracy & Loss & Accuracy & Loss & Accuracy & Loss \\ \midrule
\multirow{4}{*}{RoBERTa} & Baseline       & 28.0\%       & 9.07       & 64.5\%    & \textbf{4.91}       & 6.1\%       & 10.99       \\ 
                               & LASER          & 30.7\%      & \textbf{7.69}   & 72.5\%    & 6.44      & \textbf{6.7\%}      & \textbf{10.53}      \\ 
                               & $\text{TRAWL}_{CP}$   & \textbf{43.46\%}    & 9.25       & \textbf{73.07\%}   & 5.20      & 6.28\%      & 12.67      \\ 
                               & $\text{TRAWL}_{Tucker}$ & 41.01\%     & 8.97       & 71.47\%   & 7.55      & 6.58\%      & 12.27      \\ \midrule
\multirow{4}{*}{GPT-J}   & Baseline       & 51.84\%      & 3.52       & 75.58\%   & \textbf{4.64}       & 19.6\%      & 3.40       \\ 
                               & LASER          & 65.9\%      & 2.86       & 82.1\%    & 4.91      & 19.5\%      & \textbf{3.39}       \\ 
                               & $\text{TRAWL}_{CP}$   & \textbf{68.1\%}     & \textbf{2.78}      & \textbf{82.35\%}  & 4.65      & \textbf{20.0\%}     & 3.47       \\ 
                               & $\text{TRAWL}_{Tucker}$ & 66.3\%      & 2.94       & 79.58\%   & 4.43      & 19.7\%      & 3.44       \\ 
                               \bottomrule
\end{tabular}
}
\caption{Performance Comparison of Baseline, LASER, and TRAWL across various Datasets for RoBERTa and GPT-J models. Best performance is represented in bold.}
\label{tab:performance}
\end{table*}

TRAWL focuses on tensor compression across multiple weight matrices, distinguishing itself from existing methods \cite{sharma2023truth, saha2024compressing} by emphasizing how matrices are stacked for decomposition.

In our approach, we explore different methods of tensor formulation to effectively apply tensor decomposition across multiple weight matrices within the transformer architecture. Specifically, we examine: 

\subsubsection{One Layer at a Time}

In this setting, we stack the selected matrices for each layer individually and apply tensor decomposition layer by layer. Specifically, for each layer $l$, we form a 3-mode tensor $\mathcal{W}_l$:

\[
\mathcal{W}_l = \left[\mathbf{Q}_l, \mathbf{K}_l, \mathbf{V}_l, \mathbf{O}_l\right].
\]

Tensor decomposition is applied to each layer, denoted as \(\mathcal{W}_l\), separately. The impact on performance is evaluated layer by layer. In this process, the matrices of each layer are replaced with their decomposed and reconstructed versions. After evaluation, the original matrices are restored. This procedure is repeated for all layers. The approach is illustrated in Figure \ref{fig:TensorPerLayer}.




\subsubsection{Across the Entire Model}
\begin{figure}[h]
  \centering
  \includegraphics[width=0.65\textwidth]{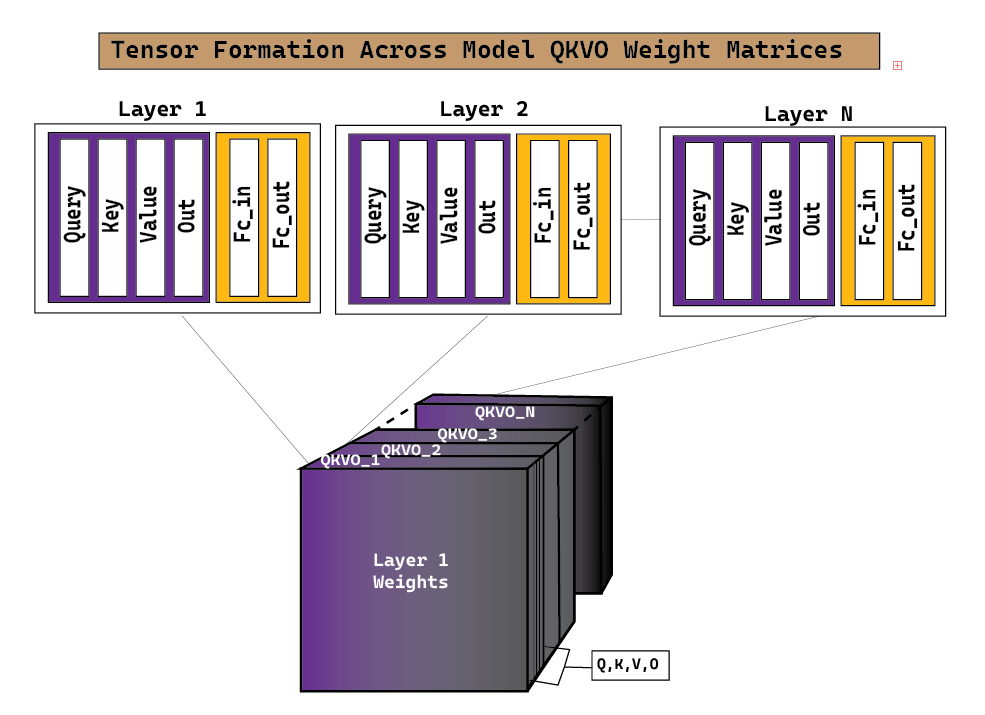}
  \caption{Tensor construction by stacking QKVO or FC weights across all layers. This forms a 3-mode tensor for tensor decomposition.}
  \label{fig:TensorAcrossModel}
\end{figure}

In this setting, we stack the selected matrices (either QKVO or FC) from all layers of the model into a single higher-order tensor. This global stacking enables tensor decomposition across the entire set of matrices, capturing interactions across all layers. Formally, let $\mathbf{W}_l$ represent the weight matrices of layer $l$, where $\mathbf{W}_l \in \{\mathbf{Q}_l, \mathbf{K}_l, \mathbf{V}_l, \mathbf{O}_l\}$ for the attention layers, or $\mathbf{W}_l$ for the FC layers. We then construct a 3-mode tensor $\mathcal{W}$:

\[
\mathcal{W} = \left[\mathbf{W}_1, \mathbf{W}_2, \dots, \mathbf{W}_L\right]
\]

where $L$ is the number of layers. Tensor decomposition is applied to $\mathcal{W}$ to capture global interactions across layers. This is illustrated in Figure \ref{fig:TensorAcrossModel}.

\subsubsection{Segmented Layers (Early, Middle, Last)}
In this configuration,  the layers are split into three segments: the initial third, intermediate third, and final third. Each segment undergoes separate tensor decomposition and performance evaluation, allowing analysis of the technique's impact across sections of the model architecture to determine which segments benefit most.

\section{Experiments and Results}

\subsection{Experimental Setups}

We primarily focus on two models, GPTJ-6B \cite{gpt-j} and RoBERTa \cite{liu2019roberta}, and tested them with three datasets, BigBench WikiQA \cite{srivastava2022beyond}, BiosProfession \cite{10.1145/3287560.3287572}, and HotPotQA \cite{yang2018hotpotqadatasetdiverseexplainable}. In our experiments, we applied two tensor decomposition methods---CP decomposition and Tucker decomposition---to better understand the effect of tensor decomposition on LLMs. These were compared with baseline experiments where no tensor decomposition was applied.


To effectively utilize tensor decomposition and low-rank approximation, we transpose one of the FC weight matrices when stacking them to form a tensor. This is necessary because the FC input and FC output matrices are transposed relative to each other in shape. By transposing, we can appropriately structure it as a suitable tensor for decomposition. Once the tensor decomposition is performed and a low-rank approximation is obtained, we transpose the resulting matrix back to its original shape. This transformed matrix then substitutes the original matrix in the LLM, ensuring that the model's architecture remains consistent while benefiting from the reduced complexity and enhanced efficiency provided by the tensor decomposition process. For the QKVO matrices, we simply stack them together to form the tensor to decompose, since they are all of the same shape, allowing for a straightforward application of tensor decomposition techniques.

\subsection{Results}

\subsubsection{One Layer at a Time}

Table~\ref{tab:performance} compares the performance of the baseline, LASER, and TRAWL methods for both RoBERTa and GPT-J models in three data sets: BigBench WikiQA, BiosProfession, and HotpotQA. 
\begin{figure*}[t]
  \includegraphics[width=0.48\linewidth]{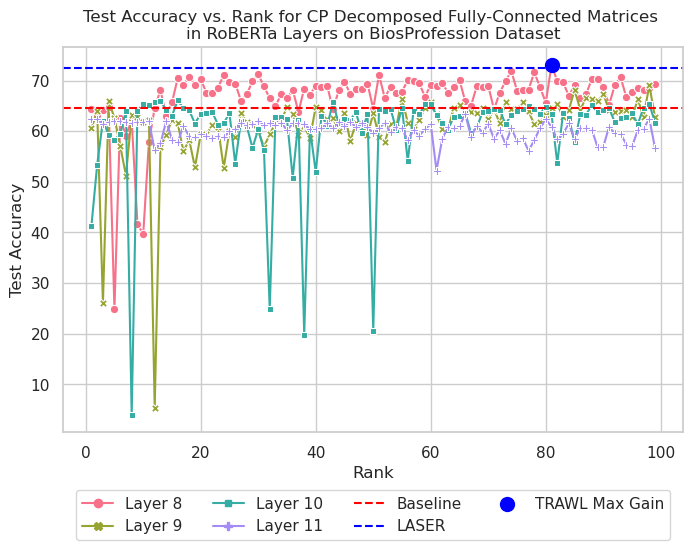} \hfill
  \includegraphics[width=0.48\linewidth]{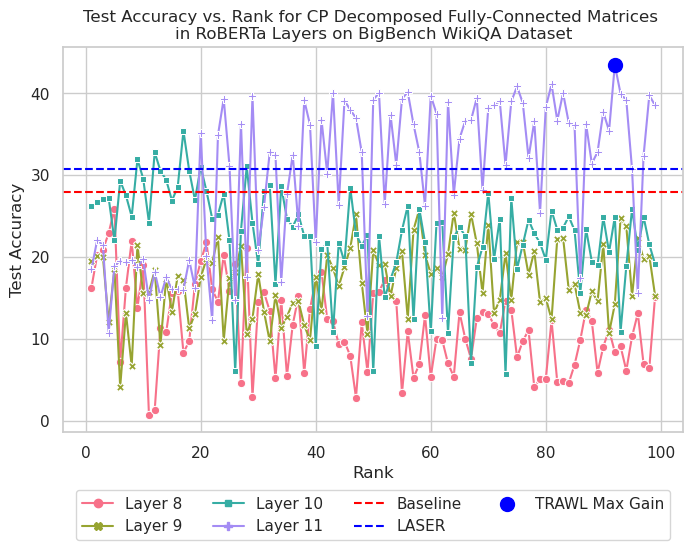}
  \caption {\textbf{RoBERTa performance when approximating the last few FC layers one at a time:} Results for the BiosProfession dataset on the left and the BigBench WikiQA dataset on the right. The blue dashed line represents the best LASER result, while the red dashed line indicates the baseline model performance without any decomposition. Decomposing the last few layers individually led to the highest performance gains.}
  \label{fig:roberta_FC_performance}
\end{figure*}

First, we observe that TRAWL, our tensor decomposition approach across multiple weight matrices, consistently outperforms the baseline compression method, LASER. For the BigBench WikiQA dataset, TRAWL$_{CP}$ delivers the best accuracy, with 43.46\% for RoBERTa and 68.1\% for GPT-J, outperforming LASER and the baseline. It also achieves the lowest loss for GPT-J, while LASER has the lowest loss for RoBERTa. In the BiosProfession dataset, TRAWL$_{CP}$ continues to lead with 73.07\% accuracy for RoBERTa and 82.35\% for GPT-J, while improving the loss of GPT-J. For HotpotQA, the improvements are smaller, with TRAWL$_{CP}$ slightly ahead of LASER for GPT-J, and similar results between methods for RoBERTa.

\textbf{Summary of results for one-layer-at-a-time: }In this setting, TRAWL consistently outperforms the baseline and LASER methods, particularly with CP decomposition in the FC layers. The most significant improvements are seen with the GPT-J model, especially on the BigBench WikiQA and BiosProfession datasets. However, the observed benefits are dependent on the data set, and HotpotQA shows only marginal improvements.

\subsubsection{Across the Entire Model}

\begin{figure}[!htp]
  \centering
  \includegraphics[width=0.6\textwidth]{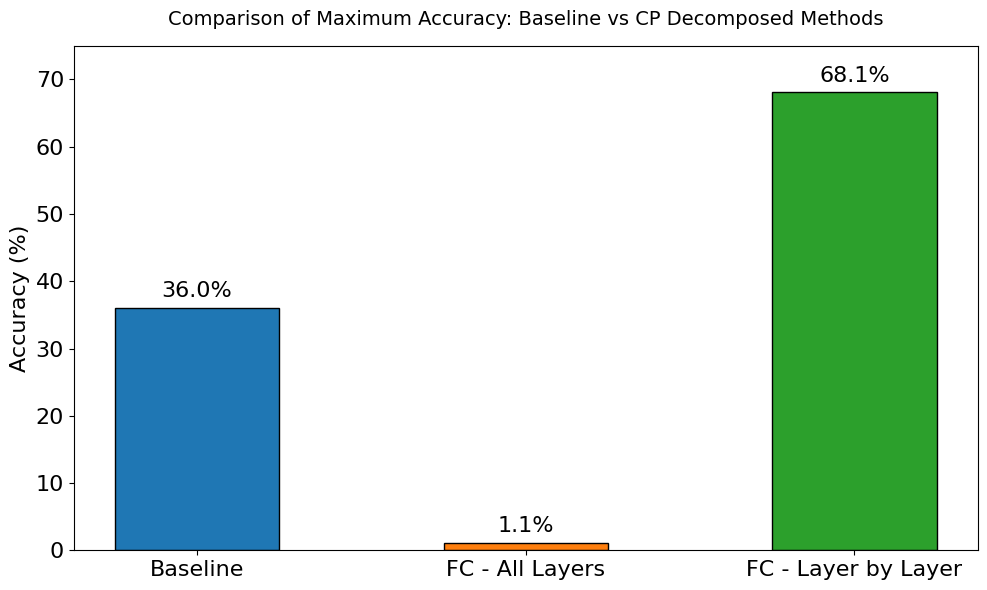}
  \caption{\textbf{Performance of CP-decomposed FC weights across the GPT-J model on BigBench WikiQA.} Combining all layers into one decomposition significantly reduces accuracy, likely due to layer heterogeneity. By contrast, a layer-by-layer approach preserves or improves performance.}
  \label{fig:FCAcrossModelWiki}
\end{figure}

When we applied CP decomposition across all FC layers of the GPT-J model, we observed a significant drop in performance. As shown in Figure \ref{fig:FCAcrossModelWiki}, the accuracy in the WikiQA dataset decreased when all layers were decomposed, a trend consistent between different ranks and components such as QKVO matrices. The combined decomposition likely introduced noise due to the varying characteristics of each layer, making it harder to capture the patterns needed to maintain performance. As a result, we switched to a layer-by-layer approach, which proved to be more effective.

\textbf{Summary of results for across-model approach: }The employed tensor formulation method did not yield promising outcomes across all datasets in the QKVO and FC cases. A possible explanation for this is that the latent patterns across all layers are widely heterogeneous and are not effectively captured within the rank budget that we set for our experiments. This may be potentially mitigated by using more complex tensor models that may be able to capture the heterogeneity more effectively. We defer this exploration to future work.

\subsubsection{Segmented Layers}
When organizing the chosen matrices into predefined sections and dividing them roughly into thirds for the initial, middle, and final parts of the model, we observed several different things. 

For RoBERTa, we have our results for this tensor formulation shown in Table \ref{tab: roberta_results_segmented_layers}. The first thing we notice is that the highest gain in accuracy occurs when decomposing the last third of the FC weights on both WikiQA and BiosProfession datasets. Besides, the early and last segments of the QKVO matrices exhibit the most declines in performance. These results underscore the importance of focusing on the final segments of the FC layers for optimal performance improvements. They also suggest that QKVO matrices are less impactful in this tensor decomposition framework.

\textbf{Summary of results for segmented layers: } This approach only produces improved accuracies when decomposing the last FC matrices.  For all the other combinations including early QKVO, middle QKVO, last QKVO, early FC, and middle FC, we get a drop in accuracy. 
Our results show that there is a smaller drop in accuracy when decomposing the middle third of the layers for QKVO weight matrices, indicating these layers are more compressible than the rest. In this type of tensor formulation, the highest gains were seen by stacking the FC matrices of the final few layers, although the gain was not as high as decomposing them one at a time.

\begin{table}[!ht]
  \centering
  \begin{tabular}{|cl|c|c|}
    \hline
    \multicolumn{2}{|c|}{}                               & WikiQA & BiosProfession  \\ \hline
    \multicolumn{1}{|c|}{\multirow{3}{*}{QKVO}} & E  & 0.17                         & 30.49                                \\ \cline{2-4} 
    \multicolumn{1}{|c|}{}                      & M & 27.11                        & 51.06                                \\ \cline{2-4} 
    \multicolumn{1}{|c|}{}                      & L  & 0.87                         & 20.00                                \\ \hline
    \multicolumn{1}{|c|}{\multirow{3}{*}{FC}}   & E  & 0.20                         & 55.16                                \\ \cline{2-4} 
    \multicolumn{1}{|c|}{}                      & M & 4.15                         & 52.19                                \\ \cline{2-4} 
    \multicolumn{1}{|c|}{}                      & L   & \textbf{31.53}               & \textbf{63.41}                       \\ \hline
    \multicolumn{2}{|c|}{Baseline Acc}                   & 27.98          & 64.57          \\ \hline
  \end{tabular}
  
  \caption{Roberta's accuracy metrics for the QKVO and FC layers across various sections of the model (\textbf{E}arly, \textbf{M}iddle, \textbf{L}ast) on the WikiQA and BiosProfession datasets. Segmenting and stacking the FC layers at the model's final third section results in the greatest performance enhancements. }
  \label{tab: roberta_results_segmented_layers}
\end{table}


\subsection{Ablations}

\begin{figure*}[!ht]
  \includegraphics[width=0.48\linewidth]{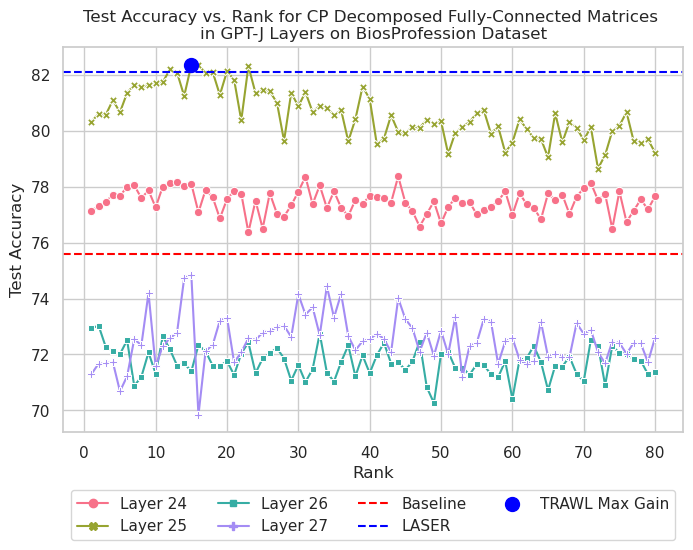} 
  \hfill
  \includegraphics[width=0.48\linewidth]{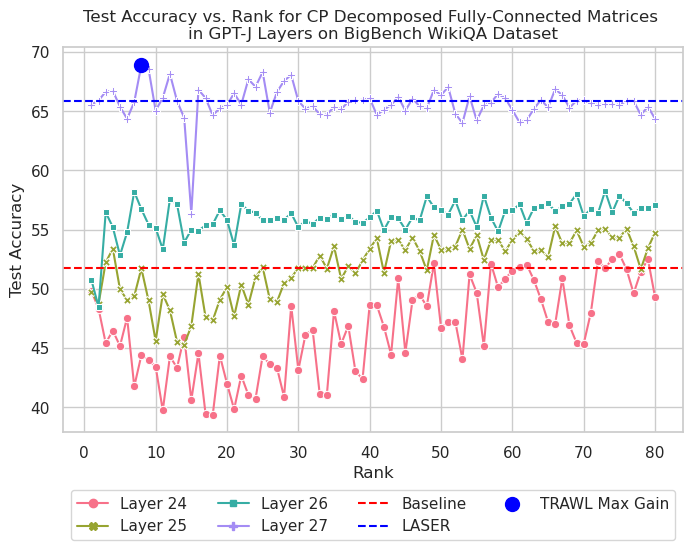}
  \caption {\textbf{GPT-J performance when approximating the last few FC layers one at a time:} Results for the BiosProfession dataset on the left and the BigBench WikiQA dataset on the right. The blue dashed line represents the best LASER result, while the red dashed line indicates the baseline model performance without any decomposition. Decomposing the last few layers individually led to the highest performance gains.}
  \label{fig:gp-ablation}
\end{figure*}

To evaluate the effect of stacking and decomposing different layers, we compared the performance of CP-decomposed fully connected (FC) weight matrices across various layers of RoBERTa on multiple datasets. The most notable improvements occurred in the FC weight matrices of the final layers, where tensor decomposition led to significant accuracy gains. In contrast, decomposing the QKVO matrices yielded minimal benefits and, in some cases, even reduced performance. As shown in Figure \ref{fig:roberta_FC_performance}, the later-stage FC layers exhibited the largest accuracy improvements, while earlier layers, particularly the QKVO matrices, showed little to no benefit, and in some instances, decreased accuracy. We also observe similar trends on GPT-J, where the results are presented in Figure \ref{fig:gp-ablation}. 

Additionally, there are more fluctuations in accuracies in RoBERTa (Figure \ref{fig:roberta_FC_performance}) compared to GPT-J, suggesting that RoBERTa's performance is more sensitive to the specific rank and decomposition applied. These findings show that while tensor decomposition is beneficial, its impact is highly dependent on the specific layer, the corresponding weight matrix, and the tensor decomposition rank.

Last but not least, the one-layer-at-a-time approach usually resulted in the highest performance gains near the final few layers when decomposing FC weights. However, no significant improvements were observed for QKVO matrices. These results align with the findings of \cite{sharma2023truth}, highlighting that FC layers are key targets for effective compression. In this work we demonstrate that there exists meaningful low-rank structure when treating the FC weights jointly as a tensor. In the future, we aim to explore whether more complex models or alternative methods can yield improvements for QKVO matrices.

\section{Conclusion}

In this study, we introduce \textbf{TRAWL}, a novel method for denoising LLMs using tensor decomposition. TRAWL extracts specific weight matrices, stacks them into a tensor, and applies decomposition to replace the original matrices with low-rank approximations. Our results show that layer-by-layer decomposition and selective layer stacking significantly improve performance, with the most notable gains seen in the fully connected layers of the terminal layers. These findings highlight the effectiveness of adaptive decomposition techniques in enhancing LLM efficiency, sometimes even outperforming original models by denoising. 

Our research presents several promising avenues for future exploration. One area that we would like to explore is the investigation of novel decomposition techniques. Alternatives like Block-Term Decomposition and randomized decomposition methods might have the potential to enhance performance even further. Additionally, enhancing the rank selection procedure is crucial, as the optimal rank plays a pivotal role in achieving optimal performance. Furthermore, we propose to extend the applicability of TRAWL by implementing it in a wider range of models and tasks. This includes, but is not limited to, domains within computer vision, natural language processing, and reinforcement learning.

\section{Limitations}

While our study focused on tensor decomposition techniques like CANDECOMP/PARAFAC (CP) and Tucker, other methods such as Tensor training or Block-Term Decomposition (BTD) may offer further performance gains. Additionally, joint compression of heterogeneous layers forming "irregular tensors" presents another promising direction to explore for enhancing model efficiency.

Estimating the optimal rank for tensor decomposition remains a challenge, as there is no universal solution like the Singular Value Decomposition for matrices. Although we explored different ranks using available heuristics, there is still no "silver bullet." Refining these rank selection methods is an important area for future research.

Our method demonstrates that compressing certain layers and weight types can significantly improve performance. However, we lack a systematic, non-brute-force approach to determine the best combination of layers and weights for compression. This is a challenge shared by prior work, such as LASER, and remains an open problem for future exploration.

Low-rank compression provides two key benefits: reducing model size through approximation and improving performance via denoising. While we focused on the latter in this study, future research should investigate the practical systems implications, such as memory and compute savings for inference on various devices.

Finally, although our findings were consistent across two language models and three benchmark datasets, the extent to which they generalize to other models and datasets requires further validation. Future work should explore the broader applicability of our approach to ensure robust performance across diverse models and tasks.

\section*{Acknowledgments}
Research was supported by the National Science Foundation under CAREER grant no. IIS 2046086, grant no. CNS 2106982, and CREST Center for Multidisciplinary Research Excellence in CyberPhysical Infrastructure Systems (MECIS) grant no. 2112650,  by the Agriculture and Food Research Initiative Competitive Grant no. 2020-69012-31914 from the USDA National Institute of Food and Agriculture, and by the University Transportation Center for Railway Safety (UTCRS) at UTRGV through the USDOT UTC Program under Grant No. 69A3552348340.

\bibliographystyle{splncs04}
\bibliography{custom.bib}

\end{document}